\documentclass[10pt,twocolumn,letterpaper]{article}

\usepackage{iccv}
\usepackage{times}
\usepackage{epsfig}
\usepackage{graphicx}
\usepackage{amsmath}
\usepackage{amssymb}
\usepackage{float}
\usepackage[toc,page,title]{appendix}
\usepackage{textcomp}
\usepackage{gensymb}
\usepackage{multirow}
\usepackage{xcolor}
\usepackage[linesnumbered,ruled]{algorithm2e}
\usepackage{balance}
\SetAlFnt{\small}
\usepackage[font=small,skip=0pt]{caption}
\DeclareMathOperator*{\argminA}{arg\,min}
\usepackage{pifont}
\usepackage[pagebackref=true,breaklinks=true,letterpaper=true,colorlinks,bookmarks=false]{hyperref}

\iccvfinalcopy 

\def\iccvPaperID{3303} 

\ificcvfinal\fi

\begin{document}

\title{ICE: Inter-instance Contrastive Encoding for Unsupervised Person Re-identification}

\author{Hao Chen\textsuperscript{1,2,3}\hskip 1em 
Benoit Lagadec\textsuperscript{3} \hskip 1em 
Francois Bremond\textsuperscript{1,2}\\
\textsuperscript{1}Inria \hskip 1em
\textsuperscript{2}Université Côte d'Azur \hskip 1em
\textsuperscript{3}European Systems Integration\\
{\tt\small \{hao.chen, francois.bremond\}@inria.fr} \hskip 1em 
{\tt\small benoit.lagadec@esifrance.net} \\
}
\maketitle
\ificcvfinal\thispagestyle{empty}\fi

\begin{abstract}
Unsupervised person re-identification (ReID) aims at learning discriminative identity features without annotations. Recently, self-supervised contrastive learning has gained increasing attention for its effectiveness in unsupervised representation learning. The main idea of instance contrastive learning is to match a same instance in different augmented views. 
However, the relationship between different instances has not been fully explored in previous contrastive methods, especially for instance-level contrastive loss.
To address this issue, we propose Inter-instance Contrastive Encoding (ICE) that leverages inter-instance pairwise similarity scores to boost previous class-level contrastive ReID methods. We first use pairwise similarity ranking as one-hot hard pseudo labels for hard instance contrast, which aims at reducing intra-class variance. Then, we use similarity scores as soft pseudo labels to enhance the consistency between augmented and original views, which makes our model more robust to augmentation perturbations.
Experiments on several large-scale person ReID datasets validate the effectiveness of our proposed unsupervised method ICE, which is competitive with even supervised methods. Code is made available at \url{https://github.com/chenhao2345/ICE}.
\end{abstract}

\section{Introduction}
Person re-identification (ReID) targets at retrieving an person of interest across non-overlapping cameras by comparing the similarity of appearance representations. Supervised ReID methods \cite{sun2018beyond,Chen_2020_WACV,Luo_2019_CVPR_Workshops} use human-annotated labels to build discriminative appearance representations which are robust to pose, camera property and view-point variation. However, annotating cross-camera identity labels is a cumbersome task, which makes supervised methods less scalable in real-world deployments. Unsupervised methods \cite{Lin2019ABC, Lin2020UnsupervisedPR, Wang_2020_CVPR} directly train a model on unlabeled data and thus have a better scalability.

Most of previous unsupervised ReID methods \cite{song2020unsupervised,ge2020mutual,zhai2020multiple} are based on unsupervised domain adaptation (UDA). UDA methods adjust a model from a labeled source domain to an unlabeled target domain. The source domain provides a good starting point that facilitates target domain adaptation. With the help of a large-scale source dataset, state-of-the-art UDA methods \cite{ge2020mutual,zhai2020multiple} significantly enhance the performance of unsupervised ReID. However, the performance of UDA methods is strongly influenced by source dataset's scale and quality. Moreover, a large-scale labeled dataset is not always available in the real world. In this case, fully unsupervised methods \cite{Lin2019ABC,Lin2020UnsupervisedPR} own more flexibility, as they do not require any identity annotation and directly learn from unlabeled data in a target domain. 

Recently, contrastive learning has shown excellent performance in unsupervised representation learning. 
State-of-the-art contrastive methods \cite{Wu2018UnsupervisedFL,chen2020simple, He_2020_CVPR} consider each image instance as a class and learns representations by matching augmented views of a same instance. As a class is usually composed of multiple positive instances, it hurts the performance of fine-grained ReID tasks when different images of a same identity are considered as different classes. Self-paced Contrastive Learning (SpCL) \cite{ge2020self} alleviates this problem by matching an instance with the centroid of the multiple positives, where each positive converges to its centroid at a uniform pace. Although SpCL has achieved impressive performance, this method does not consider inter-instance affinities, which can be leveraged to reduce intra-class variance and make clusters more compact. In supervised ReID, state-of-the-art methods \cite{Chen_2020_WACV,Luo_2019_CVPR_Workshops} usually adopt a hard triplet loss \cite{hermans2017defense} to lay more emphasis on hard samples inside a class, so that hard samples can get closer to normal samples. In this paper, we introduce Inter-instance Contrastive Encoding (ICE), in which we match an instance with its hardest positive in a mini-batch to make clusters more compact and improve pseudo label quality. Matching the hardest positive refers to using one-hot “hard” pseudo labels.

Since no ground truth is available, mining hardest positives within clusters is likely to introduce false positives into the training process. In addition, the one-hot label does not take the complex inter-instance relationship into consideration when multiple pseudo positives and negatives exist in a mini-batch. Contrastive methods usually use data augmentation to mimic real-world distortions, \eg, occlusion, view-point and resolution variance. After data augmentation operations, certain pseudo positives may become less similar to an anchor, while certain pseudo negatives may become more similar.
As a robust model should be invariant to distortions from data augmentation, we propose to use the inter-instance pairwise similarity as “soft” pseudo labels to enhance the consistency before and after augmentation. 

Our proposed ICE incorporates class-level label (centroid contrast), instance pairwise hard label (hardest positive contrast) and instance pairwise soft label (augmentation consistency) into one fully unsupervised person ReID framework. Without any identity annotation, ICE significantly outperforms state-of-the-art UDA and fully unsupervised methods on main-stream person ReID datasets.

To summarize, our contributions are: (1) We propose to use pairwise similarity ranking to mine hardest samples as one-hot hard pseudo labels for hard instance contrast, which reduces intra-class variance. (2) We propose to use pairwise similarity scores as soft pseudo labels to enhance the consistency between augmented and original instances, which alleviates label noise and makes our model more robust to augmentation perturbation. (3) Extensive experiments highlight the importance of inter-instance pairwise similarity in contrastive learning.  Our proposed method ICE outperforms state-of-the-art methods by a considerable margin, significantly pushing unsupervised ReID to real-world deployment.

\section{Related Work}
\paragraph{Unsupervised person ReID.} Recent unsupervised person ReID methods can be roughly categorized into unsupervised domain adaptation (UDA) and fully unsupervised methods. Among UDA-based methods, several works \cite{Wang2018TransferableJA, Lin2018MultitaskMF} leverage semantic attributes to reduce the domain gap between source and target domains. Several works \cite{wei2018person, Zhong_2018_ECCV, chen2019instance, zhong2019invariance, Zou2020JointDA,Chen_2021_CVPR} use generative networks to transfer labeled source domain images into the style of target domain. Another possibility is to assign pseudo labels to unlabeled images, where pseudo labels are obtained from clustering \cite{song2020unsupervised,fu2019self,zhang2019self,Chen_2021_WACV} or reference data \cite{Yu2019UnsupervisedPR}. Pseudo label noise can be reduced by selecting credible samples \cite{chen2020deep} or using a teacher network to assign soft labels \cite{ge2020mutual}.
All these UDA-based methods require a labeled source dataset. Fully unsupervised methods have a better flexibility for deployment. BUC \cite{Lin2019ABC} first treats each image as a cluster and progressively merge clusters. Lin \etal \cite{Lin2020UnsupervisedPR} replace clustering-based pseudo labels with similarity-based softened labels. Hierarchical Clustering is proposed in \cite{zeng2020hierarchical} to improve the quality of pseudo labels. 
Since each identity usually has multiple positive instances, MMCL \cite{Wang_2020_CVPR} introduces a memory-based multi-label classification loss into unsupervised ReID. JVTC \cite{li2020joint} and CycAs \cite{wang2020CycAs} explore temporal information to refine visual similarity. SpCL \cite{ge2020self} considers each cluster and outlier as a single class and then conduct instance-to-centroid contrastive learning. CAP \cite{Wang2021camawareproxies} calculates identity centroids for each camera and conducts intra- and inter-camera centroid contrastive learning. Both SpCL and CAP focus on instance-to-centroid contrast, but neglect inter-instance affinities. 
\vspace{-10pt}
\paragraph{Contrastive Learning.}
Recent contrastive learning methods \cite{Wu2018UnsupervisedFL, He_2020_CVPR, chen2020simple} consider unsupervised representation learning as a dictionary look-up problem. Wu \etal \cite{Wu2018UnsupervisedFL} retrieve a target representation from a memory bank that stores representations of all the images in a dataset. MoCo \cite{He_2020_CVPR} introduces a momentum encoder and a queue-like memory bank to dynamically update negatives for contrastive learning. In SimCLR \cite{chen2020simple}, authors directly retrieve representations within a large batch. However, all these methods consider different instances of a same class as different classes, which is not suitable in a fine-grained ReID task. These methods learn invariance from augmented views, which can be regarded as a form of consistency regularization.
\vspace{-10pt}
\paragraph{Consistency regularization.} 
Consistency regularization refers to an assumption that model predictions should be consistent when fed perturbed versions of the same image, which is widely considered in recent semi-supervised learning \cite{Tarvainen2017MeanTA,sohn2020fixmatch,chen2020big}. The perturbation can come from data augmentation \cite{sohn2020fixmatch}, temporal ensembling \cite{Tarvainen2017MeanTA,Laine2016TemporalEF, ge2020self_eccv} and shallow-deep features \cite{zheng2021rectifying,chen2020big}. Artificial perturbations are applied in contrastive learning as strong augmentation \cite{chen2020improved, wei2021co} and momentum encoder \cite{He_2020_CVPR} to make a model robust to data variance. Based on temporal ensembling, Ge \etal \cite{ge2020self_eccv} use inter-instance similarity to mitigate pseudo label noise between different training epochs for image localization.
Wei \etal \cite{wei2021co} propose to regularize inter-instance consistency between two sets of augmented views, which neglects intra-class variance problem. We simultaneously reduce intra-class variance and regularize consistency between augmented and original views, which is more suitable for fine-grained ReID tasks. 

\section{Proposed Method}
\begin{figure}
\centering
   \includegraphics[width=0.9\linewidth]{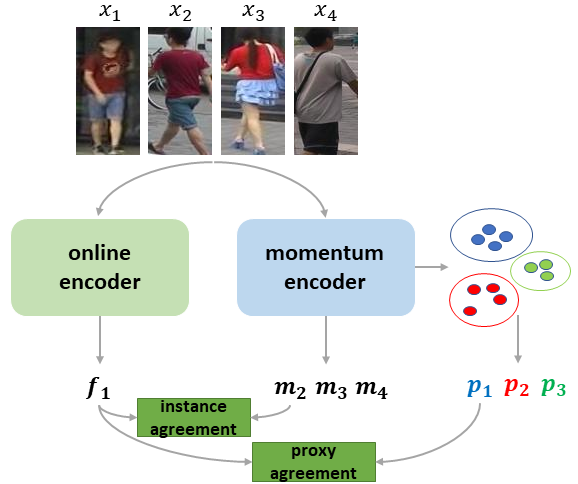}
   \caption{General architecture of ICE. We maximize the similarity between anchor and pseudo positives in both inter-class (proxy agreement between an instance representation $f_1$ and its cluster proxy $p_1$) and intra-class (instance agreement between $f_1$ and its pseudo positive $m_2$) manners. }
\label{fig:figure1}
\end{figure}
\subsection{Overview}
Given a person ReID dataset $\mathcal{X}=\{x_{1}, x_{2}, ..., x_{N}\}$, our objective is to train a robust model on $\mathcal{X}$ without annotation. For inference, representations of a same person are supposed to be as close as possible. State-of-the-art contrastive methods \cite{He_2020_CVPR,chen2020simple} consider each image as an individual class and maximize similarities between augmented views of a same instance with InfoNCE loss \cite{Oord2018RepresentationLW}:
\begin{equation}
   \mathcal{L}_{InfoNCE} = \mathop{\mathbb{E}}[-\log{\frac{\exp{(q \cdot k_{+}/\tau)}}{\sum\nolimits_{i=0}^{K}\exp{(q \cdot k_{i}/\tau)}}}]
\label{InfoNCE}
\end{equation}
where $q$ and $k_{+}$ are two augmented views of a same instance in a set of candidates $k_{i}$. $\tau$ is a temperature hyper-parameter that controls the scale of similarities. 


Following MoCo \cite{He_2020_CVPR}, we design our proposed ICE with an online encoder and a momentum encoder as shown in Fig.~\ref{fig:figure1}. The online encoder is a regular network, \eg, ResNet50 \cite{he2016deep}, which is updated by back-propagation. The momentum encoder (weights noted as $\theta_m$) has the same structure as the online encoder, but updated by accumulated weights of the online encoder (weights noted as $\theta_o$):
\begin{equation}
\theta_m^t =\alpha\theta_m^{t-1}+(1-\alpha)\theta_o^t
\label{equ:ema}
\end{equation}
where $\alpha$ is a momentum coefficient that controls the update speed of the momentum encoder. $t$ and $t-1$ refer respectively to the current and last iteration. The momentum encoder builds momentum representations with the moving averaged weights, which are more stable to label noise.

At the beginning of each training epoch, we use the momentum encoder to extract appearance representations $\mathcal{M}=\{m_{1}, m_{2}, ..., m_{N}\}$ of all the samples in the training set $\mathcal{X}$. We use a clustering algorithm DBSCAN \cite{Ester1996ADA} on these appearance representations to generate pseudo identity labels $\mathcal{Y}=\{y_{1}, y_{2}, ..., y_{N}\}$. We only consider clustered inliers for contrastive learning, while un-clustered outliers are discarded. We calculate proxy centroids $p_{1}, p_{2}, ...$ and store them in a memory for a proxy contrastive loss $\mathcal{L}_{proxy}$ (see Sec.~\ref{section: center contrastive loss}). Note that this proxy memory can be camera-agnostic \cite{ge2020self} or camera-aware \cite{Wang2021camawareproxies}.

Then, we use a random identity sampler to split the training set into mini-batches where each mini-batch contains $N_P$ pseudo identities and each identity has $N_K$ instances. We train the whole network by combining the $\mathcal{L}_{proxy}$ (with class-level labels), a hard instance contrastive loss $\mathcal{L}_{h\_ins}$ (with hard instance pairwise labels, see Sec.~\ref{section: Hard Instance Contrastive Loss}) and a soft instance consistency loss $\mathcal{L}_{s\_ins}$ (with soft instance pairwise labels, see Sec.~\ref{section: Consistency Loss}):
\begin{equation}
\mathcal{L}_{total} = \mathcal{L}_{proxy}+\lambda_{h} \mathcal{L}_{h\_ins}+\lambda_{s} \mathcal{L}_{s\_ins}
\label{equ:total}
\end{equation}

To increase the consistency before and after data augmentation, we use different augmentation settings for prediction and target representations in the three losses (see Tab.~\ref{tab:augmentation}).  
\begin{table}[H]
\centering
\scalebox{0.8}{
\begin{tabular}
{ccc}
\hline
Loss& Predictions (augmentation)& Targets (augmentation) \\
\hline
$\mathcal{L}_{proxy}$& $f$ (Strong) & $p$ (None) \\
$\mathcal{L}_{h\_ins}$& $f$ (Strong) & $m$ (Strong) \\
$\mathcal{L}_{s\_ins}$ & $P$ (Strong) & $Q$ (None) \\
\hline
\end{tabular}}
\caption{Augmentation settings for 3 losses.  }
\label{tab:augmentation}
\end{table}

\subsection{Proxy Centroid Contrastive Baseline}
\label{section: center contrastive loss}

\paragraph{For a camera-agnostic memory,} the proxy of cluster $a$ is defined as the averaged momentum representations of all the instances belonging to this cluster:
\begin{equation}
   p^{}_{a} = \frac{1}{N_{a}}\sum_{m_{i} \in y_{a}} m^{}_{i}
\label{equa:center_a}
\end{equation}
where $N_a$ is the number of instances belonging to the cluster $a$.

\begin{figure}
\centering
   \includegraphics[width=0.8\linewidth]{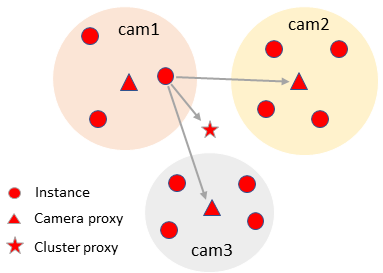}
   \caption{Proxy contrastive loss. Inside a cluster, an instance is pulled to a cluster centroid by $L_{agnostic}$ and to cross-camera centroids by $L_{cross}$. }
\label{fig:figure2}
\end{figure}

We apply a set of data augmentation on $\mathcal{X}$ and feed them to the online encoder.
For an online representation $f_{a}$ belonging to the cluster $a$, the camera-agnostic proxy contrastive loss is a softmax log loss with one positive proxy $p^{}_{a}$ and all the negatives in the memory: 
\begin{equation}
   \mathcal{L}_{agnostic} = \mathop{\mathbb{E}}[-\log{\frac{\exp{(f_{a} \cdot p_{a}/\tau_{a})}}{\sum\nolimits_{i=1}^{|p^{}|}\exp{(f_{a} \cdot p_{i}/\tau_{a})}}}]
\label{equa:centerloss_a}
\end{equation}
where $|p^{}|$ is the number of clusters in a training epoch and $\tau_{a}$ is a temperature hyper-parameter. Different from unified contrastive loss \cite{ge2020mutual}, outliers are not considered as single instance clusters. In such way, outliers are not pushed away from clustered instances, which allows us to mine more hard samples for our proposed hard instance contrast. As shown in Fig.~\ref{fig:figure2}, all the clustered instances converge to a common cluster proxy centroid. However, images inside a cluster are prone to be affected by camera styles, leading to high intra-class variance. This problem can be alleviated by adding a cross-camera proxy contrastive loss \cite{Wang2021camawareproxies}. 

\vspace{-10pt}
\paragraph{For a camera-aware memory,} if we have $\mathcal{C}=\{ c_1, c_2, ...\}$ cameras, a camera proxy $p_{ab}$ is defined as the averaged momentum representations of all the instances belonging to the cluster $a$ in camera $c_b$: 
\begin{equation}
p^{}_{ab} = \frac{1}{N_{ab}}\sum_{m_{i} \in y_{a} \cap m_{i} \in c_{b}}^{}  m^{}_{i}
\label{equa:center_c}
\end{equation}
where $N_{ab}$ is the number of instances belonging to the cluster $a$ captured by camera $c_b$.

Given an online representation $f_{ab}$, the cross-camera proxy contrastive loss is a softmax log loss with one positive cross-camera proxy $p^{}_{ai}$ and $N_{neg}$ nearest negative proxies in the memory: 
\begin{equation} \small
   \mathcal{L}_{cross} = \mathop{\mathbb{E}}[- \frac{1}{|\mathcal{P}|} \sum_{i\neq b \cap i \in \mathcal{C}} \log{\frac{\exp{(<f_{ab} \cdot p_{ai}>/\tau_{c})}}{\sum_{j=1}^{N_{neg}+1}\exp{(<f_{ab} \cdot p_{j}>/\tau_{c})}}}]
\label{equa:centerloss_c}
\end{equation}
where $<\cdot >$ denotes cosine similarity and $\tau_{c}$ is a cross-camera temperature hyper-parameter. $|\mathcal{P}|$ is the number of cross-camera positive proxies. Thanks to this cross-camera proxy contrastive loss, instances from one camera are pulled closer to proxies of other cameras, which reduces intra-class camera style variance.

We define a proxy contrastive loss by combining cluster and camera proxies with a weighting coefficient $0.5$ from \cite{Wang2021camawareproxies}:
\begin{equation}
\mathcal{L}_{proxy} = \mathcal{L}_{agnostic}+0.5 \mathcal{L}_{cross}
\label{equ:proxy}
\end{equation}



\subsection{Hard Instance Contrastive Loss}
\label{section: Hard Instance Contrastive Loss}
Although intra-class variance can be alleviated by cross-camera contrastive loss, it has two drawbacks: 1) more memory space is needed to store camera-aware proxies, 2) impossible to use when camera ids are unavailable. We propose a camera-agnostic alternative by exploring inter-instance relationship instead of using camera labels. Along with training, the encoders become more and more strong, which helps outliers progressively enter clusters and become hard inliers. Pulling hard inliers closer to normal inliers effectively increases the compactness of clusters.

A mini-batch is composed of $N_P$ identities, where each identity has $N_K$ positive instances. Given an anchor instance $f^{i}$ belonging to the $i$th class, we sample the hardest positive momentum representation $m^{i}_{k}$ that has the lowest cosine similarity with $f^{i}$, see Fig.~\ref{fig:big figure}. For the same anchor, we have $J=(N_P-1)\times N_K$ negative instances that do not belong to the $i$th class. The hard instance contrastive loss for $f^{i}$ is a softmax log loss of $J+1$ (1 positive and J negative) pairs, which is defined as: 
\begin{equation}
   \mathcal{L}_{h\_ins} = \mathop{\mathbb{E}}[-\log{\frac{\exp{(<f^{i} \cdot m^{i}_{k}>/\tau_{h\_ins})}}{\sum\nolimits_{j=1}^{J+1}\exp{(<f^{i} \cdot m^{}_{j}>/\tau_{h\_ins})}}}]
\label{equa:viewloss}
\end{equation}
where $k = \argminA_{k=1,..,N_K} (<f^{i} \cdot m^{i}_{k}>)$ and $\tau_{h\_ins}$ is the hard instance temperature hyper-parameter. 
By minimizing the distance between the anchor and the hardest positive and maximizing the distance between the anchor and all negatives, $\mathcal{L}_{h\_ins}$ increases intra-class compactness and inter-class separability.  
\begin{figure}[t]
\centering
   \includegraphics[width=0.8\linewidth]{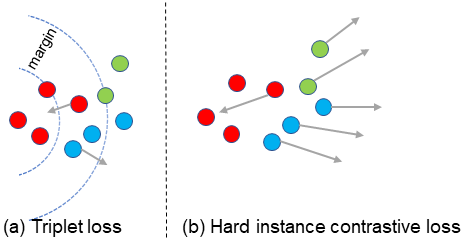}
   \caption{Comparison between triplet and hard instance contrastive loss.}
\label{fig:instance contrsative loss vs triplet loss}
\end{figure}

\paragraph{Relation with triplet loss.} Both $\mathcal{L}_{h\_ins}$ and triplet loss \cite{hermans2017defense} pull an anchor closer to positive instances and away from negative instances. As shown in Fig.~\ref{fig:instance contrsative loss vs triplet loss}, the traditional triplet loss pushes away a negative pair from a positive pair by a margin. Differently, the proposed $\mathcal{L}_{h\_ins}$ pushes away all the negative instances as far as it could with a softmax. If we select one negative instance, the $\mathcal{L}_{h\_ins}$ can be transformed into the triplet loss. If we calculate pairwise distance within a mini-batch to select the hardest positive and the hardest negative instances, the $\mathcal{L}_{h\_ins}$ is equivalent to the batch-hard triplet loss\cite{hermans2017defense}. We compare hard triplet loss (hardest negative) with the proposed $\mathcal{L}_{h\_ins}$ (all negatives). in Tab.~\ref{table:vcl}.

\begin{figure*}
\centering
  \includegraphics[width=0.9\linewidth]{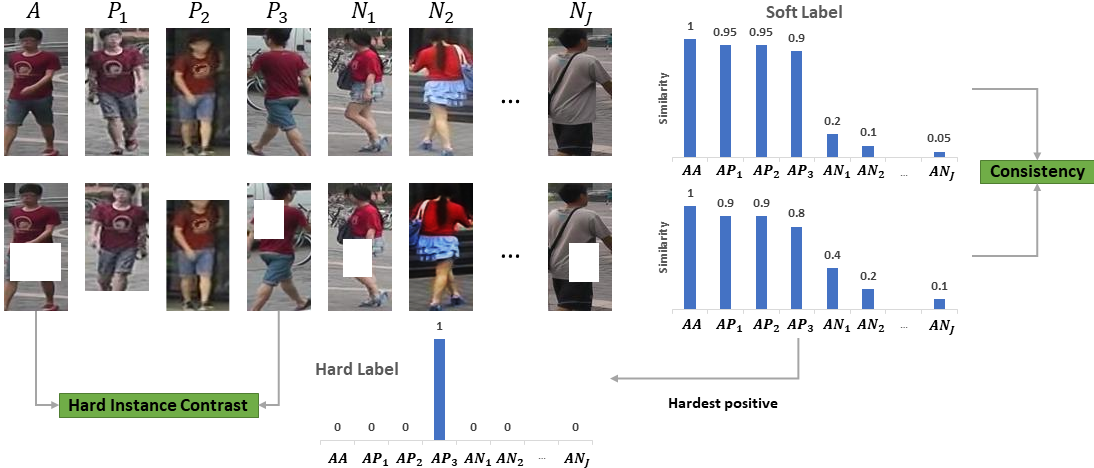}
  \caption{Based on inter-instance similarity ranking between anchor (A), pseudo positives (P) and pseudo negatives (N), \textbf{Hard Instance Contrastive Loss} matches an anchor with its hardest positive in a mini-batch. \textbf{Soft Instance Consistency Loss} regularizes the inter-instance similarity before and after data augmentation.\vspace{-0.3cm}}
\label{fig:big figure}
\end{figure*}

\begin{table}[ht]
\scalebox{0.8}{
\begin{tabular}{c|cc|cc}
\hline
\multirow{2}{*}{Negative in $\mathcal{L}_{h\_ins}$}  & \multicolumn{2}{c}{Market1501} & \multicolumn{2}{|c}{DukeMTMC-reID} \\ \cline{2-5}
\multicolumn{1}{c|}{} & \multicolumn{1}{c}{mAP} & \multicolumn{1}{c|}{Rank1} & \multicolumn{1}{c}{mAP} & \multicolumn{1}{c}{Rank1} \\ \hline
hardest&80.1&92.8&68.2&82.5\\
all&\textbf{82.3}&\textbf{93.8}&\textbf{69.9}&\textbf{83.3}\\
\hline
\end{tabular}}
\centering
\caption{Comparison between using the hardest negative and all negatives in the denominator of $\mathcal{L}_{h\_ins}$. }
\label{table:vcl}
\end{table}
\vspace{-2mm}

\subsection{Soft Instance Consistency Loss}
\label{section: Consistency Loss}
Both proxy and hard instance contrastive losses are trained with one-hot hard pseudo labels, which can not capture the complex inter-instance similarity relationship between multiple pseudo positives and negatives. Especially, inter-instance similarity may change after data augmentation. As shown in Fig.~\ref{fig:big figure}, the anchor $A$ becomes less similar to pseudo positives ($P_1, P_2, P_3$), because of the visual distortions. Meanwhile, the anchor $A$ becomes more similar to pseudo negatives ($N_1, N_2$), since both of them have red shirts. By maintaining the consistency before and after augmentation, a model is supposed to be more invariant to augmentation perturbations. We use the inter-instance similarity scores without augmentation as soft labels to rectify those with augmentation. 

For a batch of images after data augmentation, we measure the inter-instance similarity between an anchor $f_A$ with all the mini-batch $N_K \times N_P$ instances, as shown in Fig.~\ref{fig:big figure}. Then, the inter-instance similarity is turned into a prediction distribution $P$ by a softmax:
\begin{equation}
   P = \frac{\exp{(<f_{A} \cdot m^{}_{}>/\tau_{s\_ins})}}{\sum\nolimits_{j=1}^{N_P\times N_K}\exp{(<f_{A} \cdot m^{}_{j}>/\tau_{s\_ins})}}
\label{equa:P}
\end{equation}
where $\tau_{s\_ins}$ is the soft instance temperature hyper-parameter. $f_A$ is an online representation of the anchor, while $m$ is momentum representation of each instance in a mini-batch.

For the same batch without data augmentation, we measure the inter-instance similarity between momentum representations of the same anchor with all the mini-batch $N_K \times N_P$ instances, because the momentum encoder is more stable. We get a target distribution $Q$:
\begin{equation}
   Q = \frac{\exp{(<m_{A} \cdot m^{}_{}>/\tau_{s\_ins})}}{\sum\nolimits_{j=1}^{N_P\times N_K}\exp{(<m_{A} \cdot m^{}_{j}>/\tau_{s\_ins})}}
\label{equa:Q}
\end{equation}

The soft instance consistency loss is Kullback-Leibler Divergence between two distributions:
\begin{equation}
   \mathcal{L}_{s\_ins} = \mathcal{D}_{KL}(P||Q)
\label{equa:consistency}
\end{equation}

In previous methods, consistency is regularized between weakly augmented and strongly augmented images \cite{sohn2020fixmatch} or two sets of differently strong augmented images \cite{wei2021co}. Some methods \cite{Laine2016TemporalEF,Tarvainen2017MeanTA} also adopted mean square error (MSE) as their consistency loss function. We compare our setting with other possible settings in Tab.~\ref{table:consistency}. 

\begin{table}[ht]
\scalebox{0.8}{
\begin{tabular}{c|cc|cc}
\hline
\multirow{2}{*}{Consistency}  & \multicolumn{2}{c}{Market1501} & \multicolumn{2}{|c}{DukeMTMC-reID} \\ \cline{2-5}
\multicolumn{1}{c|}{} & \multicolumn{1}{c}{mAP} & \multicolumn{1}{c|}{Rank1} & \multicolumn{1}{c}{mAP} & \multicolumn{1}{c}{Rank1} \\ \hline
MSE&80.0&92.7&68.4&82.1\\
Strong-strong Aug&80.4&92.8&68.2&82.5\\
ours&\textbf{82.3}&\textbf{93.8}&\textbf{69.9}&\textbf{83.3}\\
\hline
\end{tabular}}
\centering
\caption{Comparison of consistency loss. Ours refers to KL divergence between images with and without data augmentation.}
\label{table:consistency}
\end{table}
\vspace{-3mm}
\begin{figure*}
\centering
  \includegraphics[width=0.85\linewidth]{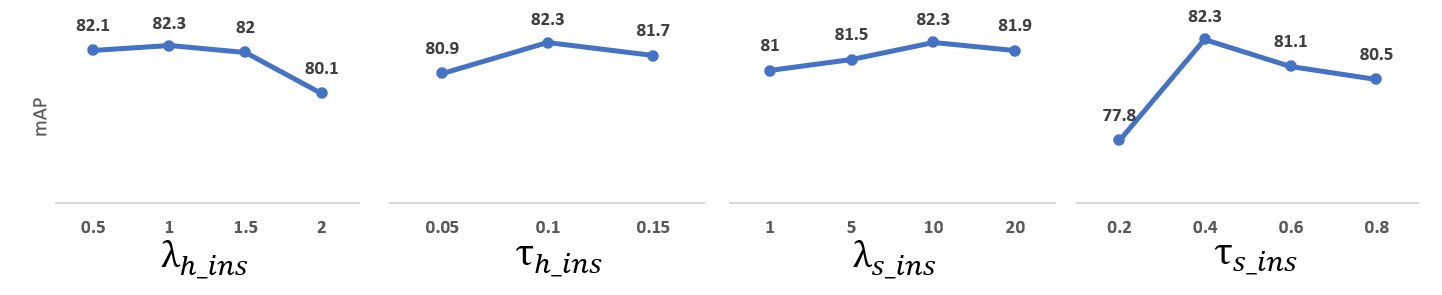}
  \caption{Parameter analysis on Market-1501 dataset. \vspace{-3mm}}
\label{fig:figure4}
\end{figure*}

\section{Experiments}
\subsection{Datasets and Evaluation Protocols}
Market-1501 \cite{Zheng2015ScalablePR}, DukeMTMC-reID\cite{ristani2016MTMC} and MSMT17 \cite{wei2018person} datasets are used to evaluate our proposed method. Market-1501 dataset is collected in front of a supermarket in Tsinghua University from 6 cameras. It contains 12,936 images of 751 identities for training and 19,732 images of 750 identities for test. DukeMTMC-reID is a subset of the DukeMTMC dataset. It contains 16,522 images of 702 persons for training, 2,228 query images and 17,661 gallery images of 702 persons for test from 8 cameras. MSMT17 is a large-scale Re-ID dataset, which contains 32,621 training images of 1,041 identities and 93,820 testing images of 3,060 identities collected from 15 cameras. Both Cumulative Matching Characteristics (CMC) Rank1, Rank5, Rank10 accuracies and mean Average Precision (mAP) are used in our experiments. 

\subsection{Implementation details}
\paragraph{General training settings.} To conduct a fair comparison with state-of-the-art methods, we use an ImageNet \cite{Russakovsky2015ImageNetLS} pre-trained ResNet50 \cite{he2016deep} as our backbone network. We report results of IBN-ResNet50 \cite{pan2018two} in Appendix~\ref{Appendix: Backbone Network}.
An Adam optimizer with a weight decay rate of 0.0005 is used to optimize our networks. The learning rate is set to 0.00035 with a warm-up scheme in the first 10 epochs. No learning rate decay is used in the training. The momentum encoder is updated with a momentum coefficient $\alpha=0.999$. We renew pseudo labels every 400 iterations and repeat this process for 40 epochs. We use a batchsize of 32 where $N_P=8$ and $N_K=4$. We set $\tau_{a}=0.5$, $\tau_{c}=0.07$ and $N_{neg}=50$ in the proxy contrastive baseline. Our network is trained on 4 Nvidia 1080 GPUs under Pytorch framework. The total training time is around 2 hours on Market-1501. After training, only the momentum encoder is used for the inference. 
\vspace{-10pt}
\paragraph{Clustering settings.} We calculate $k$-reciprocal Jaccard distance \cite{zhong2017re} for clustering, where $k$ is set to 30. We set a minimum cluster samples to 4 and a distance threshold to 0.55 for DBSCAN. We also report results of a smaller threshold 0.5 (more appropriate for the smaller dataset Market1501) and a larger threshold 0.6 (more appropriate for the larger dataset MSMT17) in Appendix~\ref{Appendix: Threshold in clustering}. 
\vspace{-10pt}
\paragraph{Data augmentation.} All images are resized to 256$\times$128. The strong data augmentation refers to random horizontal flipping, cropping, Gaussian blurring and erasing \cite{Zhong2020RandomED}.


\subsection{Parameter analysis}
Compared to the proxy contrastive baseline, ICE brings in four more hyper-parameters, including $\lambda_{h\_ins}$, $\tau_{h\_ins}$ for hard instance contrastive loss and $\lambda_{s\_ins}$, $\tau_{s\_ins}$ for soft instance consistency loss. We analyze the sensitivity of each hyper-parameter on the Market-1501 dataset. The mAP results are illustrated in Fig.~\ref{fig:figure4}. As hardest positives are likely to be false positives, an overlarge $\lambda_{h\_ins}$ or undersized $\tau_{h\_ins}$ introduce more noise. $\lambda_{h\_ins}$ and $\lambda_{s\_ins}$ balance the weight of each loss in Eq.~(\ref{equ:total}). Given the results, we set $\lambda_{h\_ins}=1$ and $\lambda_{s\_ins}=10$. $\tau_{h\_ins}$ and $\tau_{s\_ins}$ control the similarity scale in hard instance contrastive loss and soft instance consistency loss. We finally set $\tau_{h\_ins}=0.1$ and $\tau_{s\_ins}=0.4$. Our hyper-parameters
are tuned on Market-1501 and kept same for DukeMTMC-reID and MSMT17. Achieving state-of-the-art results simultaneously on the three datasets can validate the generalizability of these hyper-parameters.

\begin{table*}
\scalebox{0.81}{
\begin{tabular}{c|cccc|cccc|cccc}
\hline
\multirow{2}{*}{Camera-aware memory}  & \multicolumn{4}{c}{Market1501} & \multicolumn{4}{|c}{DukeMTMC-reID}& \multicolumn{4}{|c}{MSMT17} \\ \cline{2-13}
\multicolumn{1}{c|}{} & \multicolumn{1}{c}{mAP} & \multicolumn{1}{c}{R1} & \multicolumn{1}{c}{R5} & \multicolumn{1}{c|}{R10} & \multicolumn{1}{c}{mAP} & \multicolumn{1}{c}{R1} &\multicolumn{1}{c}{R5} & \multicolumn{1}{c|}{R10} & \multicolumn{1}{c}{mAP} & \multicolumn{1}{c}{R1} &\multicolumn{1}{c}{R5} &\multicolumn{1}{c}{R10} \\ \hline
Baseline $\mathcal{L}_{proxy}$&79.3&91.5&96.8&97.6&67.3&81.4&90.8&92.9&36.4&67.8&78.7&82.5\\
$+\mathcal{L}_{h\_ins}$&80.5&92.6&97.3&98.4&68.8&82.4&90.4&93.6&38.0&69.1&79.9&83.4\\
$+\mathcal{L}_{s\_ins}$&81.1&93.2&97.5&98.5&68.4&82.0&91.0&93.2&38.1&68.7&79.8&83.7\\
$+\mathcal{L}_{h\_ins}+\mathcal{L}_{s\_ins}$&\textbf{82.3}&\textbf{93.8}&\textbf{97.6}&\textbf{98.4}&\textbf{69.9}&\textbf{83.3}&\textbf{91.5}&\textbf{94.1}&\textbf{38.9}&\textbf{70.2}&\textbf{80.5}&\textbf{84.4}\\
\hline
\hline
\multirow{2}{*}{Camera-agnostic memory}  & \multicolumn{4}{c}{Market1501} & \multicolumn{4}{|c}{DukeMTMC-reID}& \multicolumn{4}{|c}{MSMT17} \\ \cline{2-13}
\multicolumn{1}{c|}{} & \multicolumn{1}{c}{mAP} & \multicolumn{1}{c}{R1} & \multicolumn{1}{c}{R5} & \multicolumn{1}{c|}{R10} & \multicolumn{1}{c}{mAP} & \multicolumn{1}{c}{R1} &\multicolumn{1}{c}{R5} & \multicolumn{1}{c|}{R10} & \multicolumn{1}{c}{mAP} & \multicolumn{1}{c}{R1} &\multicolumn{1}{c}{R5} &\multicolumn{1}{c}{R10} \\ \hline
Baseline $\mathcal{L}_{agnostic}$ &65.8&85.3&95.1&96.6&50.9&67.9&81.6&86.6&24.1&52.3&66.2&71.6\\
$+\mathcal{L}_{h\_ins}$&78.2&91.3&96.9&98.0&65.4&79.6&88.9&91.9&\textbf{30.3}&\textbf{60.8}&\textbf{72.9}&\textbf{77.6}\\
$+\mathcal{L}_{s\_ins}$&47.2&66.7&86.0&91.6&36.2&50.4&70.3&76.3&17.8&38.8&54.2&60.9\\
$+\mathcal{L}_{h\_ins}+\mathcal{L}_{s\_ins}$&\textbf{79.5}&\textbf{92.0}&\textbf{97.0}&\textbf{98.1}&\textbf{67.2}&\textbf{81.3}&\textbf{90.1}&\textbf{93.0}&29.8&59.0&71.7&77.0\\
\hline
\end{tabular}}
\centering
\caption{Comparison of different losses. Camera-aware memory occupies up to 6, 8 and 15 times memory space than camera-agnostic memory on Market1501, DukeMTMC-reID and MSMT17 datasets. \vspace{-0.3cm}}
\label{table:losses}
\end{table*}

\begin{figure}
\centering
   \includegraphics[width=0.8\linewidth]{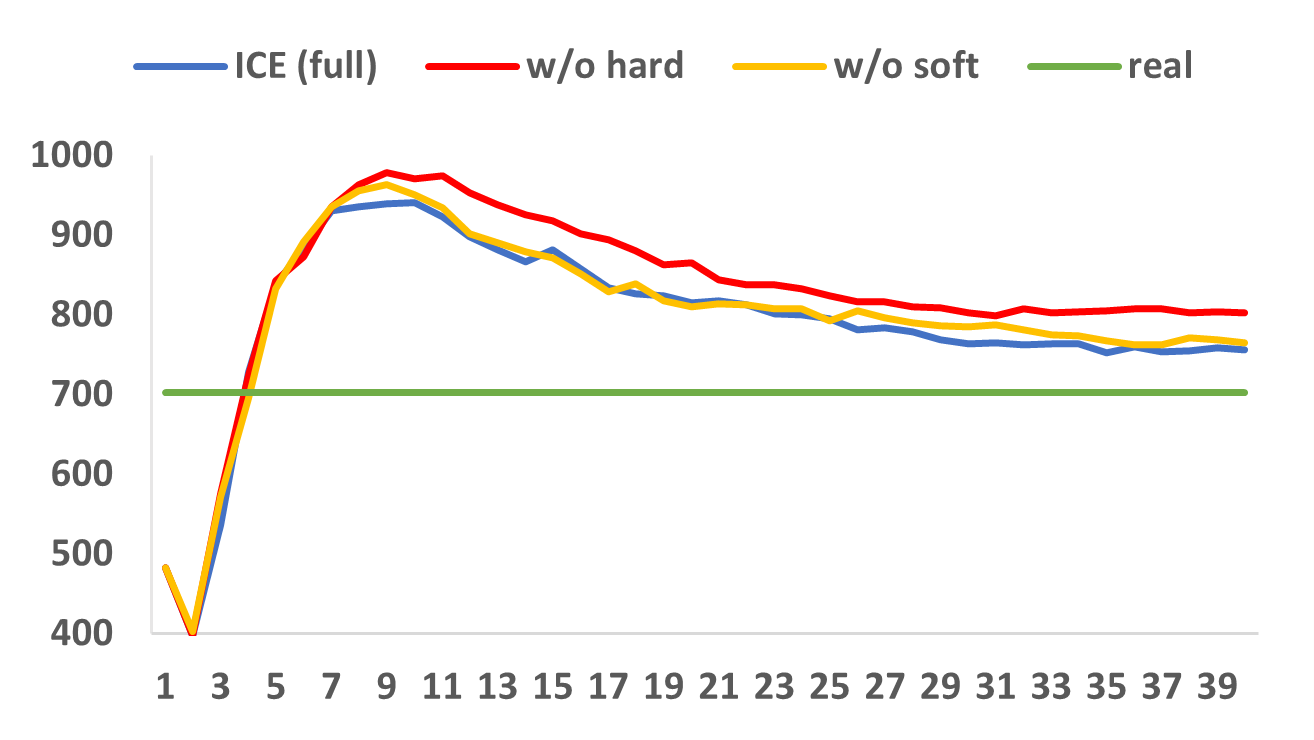}
   \caption{Dynamic cluster numbers during 40 training epochs on DukeMTMC-reID. ``hard" and ``soft" respectively denote $L_{h\_ins}$ and $L_{s\_ins}$. A lower number denotes that clusters are more compact. \vspace{-2mm}}
\label{fig:Dynamic cluster numbers}
\end{figure}

\begin{figure}
\centering
   \includegraphics[width=0.8\linewidth]{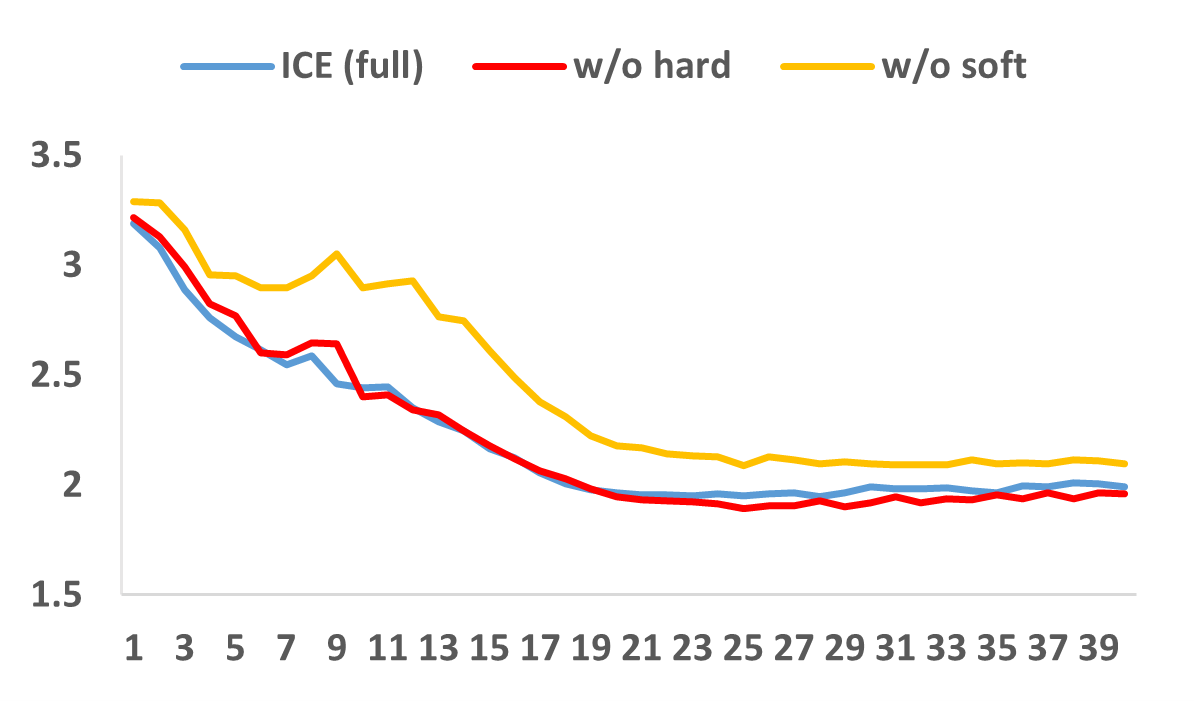}
   \caption{Dynamic KL divergence during 40 training epochs on DukeMTMC-reID. Lower KL divergence denotes that a model is more robust to augmentation perturbation.}
\label{fig:Dynamic KL divergence}
\end{figure}

\subsection{Ablation study}
\begin{figure}
\centering
   \includegraphics[width=1\linewidth]{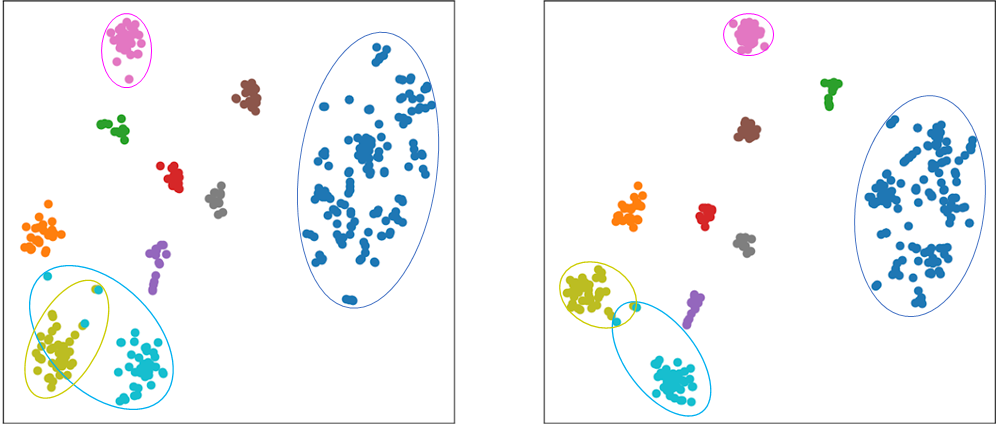}
   \caption{T-SNE visualization of 10 random classes in DukeMTMC-reID test set between \textbf{camera-aware baseline (Left)} and \textbf{ICE (Right)}.}
\label{fig:TSNE}
\end{figure}

The performance boost of ICE in unsupervised ReID mainly comes from the proposed hard instance contrastive loss and soft instance consistency loss. We conduct ablation experiments to validate the effectiveness of each loss, which is reported in Tab.~\ref{table:losses}. We illustrate the number of clusters during the training in Fig.~\ref{fig:Dynamic cluster numbers} and t-SNE \cite{vanDerMaaten2008} after training in Fig.~\ref{fig:TSNE} to evaluate the compactness of clusters. We also illustrate the dynamic KL divergence of Eq.~(\ref{equa:consistency}) to measure representation sensitivity to augmentation perturbation in Fig.~\ref{fig:Dynamic KL divergence} .   
\paragraph{Hard instance contrastive loss.} Our proposed $\mathcal{L}_{h\_ins}$ reduces the intra-class variance in a camera-agnostic manner, which increases the quality of pseudo labels. By reducing intra-class variance, a cluster is supposed to be more compact. With a same clustering algorithm, we expect to have less clusters when clusters are more compact. As shown in Fig.~\ref{fig:Dynamic cluster numbers}, DBSCAN generated more clusters during the training without our proposed $\mathcal{L}_{h\_ins}$. The full ICE framework has less clusters, which are closer to the real number of identities in the training set. On the other hand, as shown in Fig.~\ref{fig:TSNE}, the full ICE framework has a better intra-class compactness and inter-class separability than the camera-aware baseline in the test set. The compactness contributes to better unsupervised ReID performance in Tab.~\ref{table:losses}.
\vspace{-10pt}
\paragraph{Soft instance consistency loss.} Hard instance contrastive loss reduces the intra-class variance between naturally captured views, while soft instance consistency loss mainly reduces the variance from artificially augmented perturbation. 
If we compare the blue (ICE full) and yellow (w/o soft) curves in Fig.~\ref{fig:Dynamic KL divergence}, we can find that the model trained without $\mathcal{L}_{s\_ins}$ is less robust to augmentation perturbation. The quantitative results in Tab.~\ref{table:losses} confirms that the $\mathcal{L}_{s\_ins}$ improves the performance of baseline. The best performance can be obtained by applying $\mathcal{L}_{h\_ins}$ and $\mathcal{L}_{s\_ins}$ on the camera-aware baseline.
\vspace{-10pt}
\paragraph{Camera-agnostic scenario.} Above results are obtained with a camera-aware memory, which strongly relies on ground truth camera ids. We further validate the effectiveness of the two proposed losses with a camera-agnostic memory, whose results are also reported in Tab.~\ref{table:losses}. Our proposed $\mathcal{L}_{h\_ins}$ significantly improves the performance from the camera-agnostic baseline. However, \textbf{$\mathcal{L}_{s\_ins}$ should be used under low intra-class variance, which can be achieved by the variance constraints on camera styles $\mathcal{L}_{cross}$ and hard samples $\mathcal{L}_{h\_ins}$}. $\mathcal{L}_{h\_ins}$ reduces intra-class variance, so that $AA\approx AP_{1}\approx AP_{2}\approx AP_{3}\approx 1$ before augmentation in Fig.~\ref{fig:big figure}. $\mathcal{L}_{s\_ins}$ permits that we still have $AA\approx AP_{1}\approx AP_{2}\approx AP_{3}\approx 1$ after augmentation. However, when strong variance exists, \eg, $AA\not\approx AP_{1}\not\approx AP_{2}\not\approx AP_{3}\not\approx 1$, maintaining this relationship equals maintaining intra-class variance, which decreases the ReID performance. 
On medium datasets (\eg, Market1501 and DukeMTMC-reID) without strong camera variance, our proposed camera-agnostic intra-class variance constraint $\mathcal{L}_{h\_ins}$ is enough to make $\mathcal{L}_{s\_ins}$ beneficial to ReID.
On large datasets (\eg, 15 cameras in MSMT17) with strong camera variance, only camera-agnostic variance constraint $\mathcal{L}_{h\_ins}$ is not enough. We provide the dynamic cluster numbers of camera-agnostic ICE in Appendix~\ref{Appendix: Camera-agnostic scenario}.



\begin{table*}
\scalebox{0.81}{
\begin{tabular}{lc|cccc|cccc|cccc}
\hline
\multirow{2}{*}{Method} & \multirow{2}{*}{Reference} & \multicolumn{4}{c}{Market1501} & \multicolumn{4}{|c}{DukeMTMC-reID} & \multicolumn{4}{|c}{MSMT17} \\ \cline{3-14}
\multicolumn{1}{c}{} &\multicolumn{1}{c|}{}& \multicolumn{1}{c}{mAP} & \multicolumn{1}{c}{R1} & \multicolumn{1}{c}{R5} & \multicolumn{1}{c|}{R10} & \multicolumn{1}{c}{mAP} & \multicolumn{1}{c}{R1}  & \multicolumn{1}{c}{R5} & \multicolumn{1}{c|}{R10}& \multicolumn{1}{c}{mAP} & \multicolumn{1}{c}{R1}  & \multicolumn{1}{c}{R5} & \multicolumn{1}{c}{R10}\\ 
\hline
\multicolumn{2}{l}{\textbf{Unsupervised Domain Adaptation}}\\
MMCL \cite{Wang_2020_CVPR} &CVPR'20 &60.4&84.4&92.8&95.0&51.4&72.4&82.9&85.0&16.2&43.6&54.3&58.9\\
JVTC \cite{li2020joint}&ECCV'20&61.1&83.8&93.0&95.2&56.2&75.0&85.1&88.2&20.3&45.4&58.4&64.3\\
DG-Net++ \cite{Zou2020JointDA}&ECCV'20&61.7&82.1&90.2&92.7&63.8&78.9&87.8&90.4&22.1&48.8&60.9&65.9\\
ECN+ \cite{zhong2020learning}&TPAMI'20&63.8&84.1&92.8&95.4&54.4&74.0&83.7&87.4&16.0&42.5&55.9&61.5\\
MMT \cite{ge2020mutual}&ICLR'20&71.2&87.7&94.9&96.9&65.1&78.0&88.8&92.5&23.3&50.1&63.9&69.8\\
DCML \cite{chen2020deep}&ECCV'20&72.6&87.9&95.0&96.7&63.3&79.1&87.2&89.4&-&-&-&-\\
MEB \cite{zhai2020multiple} & ECCV'20&76.0&89.9&96.0&97.5&66.1& 79.6& 88.3& 92.2&-&-&-&-\\
SpCL \cite{ge2020self}&NeurIPS'20 &76.7&90.3&96.2&97.7&68.8&\textcolor{blue}{82.9}&\textcolor{blue}{90.1}&92.5&26.8&53.7&65.0&69.8\\
ABMT \cite{Chen_2021_WACV}&WACV'21 &78.3&\textcolor{blue}{92.5}&-&-&\textcolor{blue}{69.1}&82.0&-&-&26.5&54.3&-&-\\
\hline
\textbf{Fully Unsupervised}\\
BUC \cite{Lin2019ABC} &AAAI'19&29.6&61.9&73.5&78.2&22.1&40.4&52.5&58.2&-&-&-&-\\ 
SSL \cite{Lin2020UnsupervisedPR}&CVPR'20&37.8&71.7&83.8&87.4&28.6&52.5&63.5&68.9&-&-&-&-\\
JVTC \cite{li2020joint}&ECCV'20&41.8&72.9&84.2&88.7&42.2&67.6&78.0&81.6&15.1&39.0&50.9&56.8\\
MMCL \cite{Wang_2020_CVPR} &CVPR'20 &45.5&80.3&89.4&92.3&40.2&65.2&75.9&80.0&11.2&35.4&44.8&49.8\\
HCT \cite{zeng2020hierarchical}&CVPR'20 &56.4&80.0&91.6&95.2&50.7&69.6&83.4&87.4&-&-&-&-\\
CycAs \cite{wang2020CycAs}&ECCV'20& 64.8&84.8&-&-&60.1&77.9&-&-&26.7&50.1&-&-\\
GCL \cite{Chen_2021_CVPR}&CVPR'21&66.8&87.3&93.5&95.5&62.8&82.9&87.1&88.5&21.3&45.7&58.6&64.5\\
SpCL(agnostic) \cite{ge2020self}&NeurIPS'20 &73.1&88.1&95.1&97.0&65.3&81.2&90.3&92.2&19.1&42.3&55.6&61.2\\
\textbf{ICE(agnostic)}&This paper&\textcolor{blue}{79.5}&92.0&\textcolor{blue}{97.0}&\textcolor{blue}{98.1}&67.2&81.3&90.1&\textcolor{blue}{93.0}&29.8&59.0&71.7&77.0\\
CAP(aware)\cite{Wang2021camawareproxies}&AAAI'21 &79.2&91.4&96.3&97.7&67.3&81.1&89.3&91.8&\textcolor{blue}{36.9}&\textcolor{blue}{67.4}&\textcolor{blue}{78.0}&\textcolor{blue}{81.4}\\
\textbf{ICE(aware)} &This paper&\textcolor{red}{82.3}&\textcolor{red}{93.8}&\textcolor{red}{97.6}&\textcolor{red}{98.4}&\textcolor{red}{69.9}&\textcolor{red}{83.3}&\textcolor{red}{91.5}&\textcolor{red}{94.1}&\textcolor{red}{38.9}&\textcolor{red}{70.2}&\textcolor{red}{80.5}&\textcolor{red}{84.4}\\
\hline
\textbf{Supervised}\\
PCB \cite{sun2018beyond}&ECCV'18&81.6&93.8&97.5&98.5&69.2&83.3&90.5&92.5&40.4&68.2&-&-\\
DG-Net \cite{zheng2019joint}&CVPR'19&86.0&94.8&-&-&74.8&86.6&-&-&52.3&77.2&-&-\\
\textbf{ICE} (w/ ground truth) &This paper&86.6&95.1&98.3&98.9&76.5&88.2&94.1&95.7&50.4&76.4&86.6&90.0\\
\hline
\end{tabular}}
\centering
\caption{Comparison of ReID methods on Market1501, DukeMTMC-reID and MSMT17 datasets. The best and second best unsupervised results are marked in \textcolor{red}{red} and \textcolor{blue}{blue}. \vspace{-2mm}}
\label{table:market duke}
\end{table*}

\subsection{Comparison with state-of-the-art methods}
We compare ICE with state-of-the-art ReID methods in Tab.~\ref{table:market duke}. 
\vspace{-2mm}
\paragraph{Comparison with unsupervised method.} Previous unsupervised methods can be categorized into unsupervised domain adaptation (UDA) and fully unsupervised methods. We first list state-of-the-art UDA methods, including MMCL \cite{Wang_2020_CVPR}, JVTC \cite{li2020joint},  DG-Net++ \cite{Zou2020JointDA}, ECN+ \cite{zhong2020learning}, MMT \cite{ge2020mutual}, DCML \cite{chen2020deep}, MEB \cite{zhai2020multiple}, SpCL \cite{ge2020self} and ABMT~\cite{Chen_2021_WACV}. UDA methods usually rely on source domain annotation to reduce the pseudo label noise. Without any identity annotation, our proposed ICE outperforms all of them on the three datasets. 

\begin{figure}
\centering
  \includegraphics[width=1\linewidth]{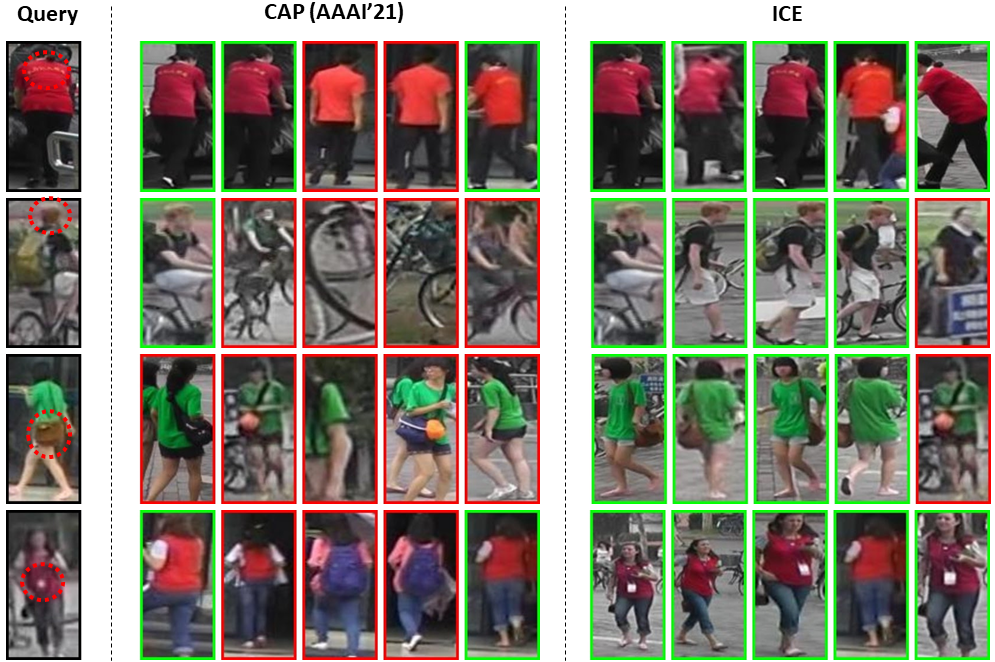}
  \caption{Comparison of top 5 retrieved images on Market1501 between CAP \cite{Wang2021camawareproxies} and ICE. Green boxes denote correct results, while red boxes denote false results. Important visual clues are marked with red dashes. }
\label{fig:Qualitative comparison}
\end{figure}

Under the fully unsupervised setting, ICE also achieves better performance than state-of-the-art methods, including BUC \cite{Lin2019ABC}, SSL \cite{Lin2020UnsupervisedPR}, MMCL \cite{Wang_2020_CVPR}, JVTC \cite{li2020joint}, HCT \cite{zeng2020hierarchical}, CycAs \cite{wang2020CycAs}, GCL~\cite{Chen_2021_CVPR}, SpCL \cite{ge2020self} and CAP \cite{Wang2021camawareproxies}. CycAs  leveraged temporal information to assist visual matching, while our method only considers visual similarity. SpCL and CAP are based on proxy contrastive learning, which are considered respectively as camera-agnostic and camera-aware baselines in our method. With a camera-agnostic memory, the performance of ICE(agnostic) remarkably surpasses the camera-agnostic baseline SpCL, especially on Market1501 and MSMT17 datasets. With a camera-aware memory, ICE(aware) outperforms the camera-aware baseline CAP on all the three datasets. By mining hard positives to reduce intra-class variance, ICE is more robust to hard samples. We illustrate some hard examples in Fig.~\ref{fig:Qualitative comparison}, where ICE succeeds to notice important visual clues, \eg, characters in the shirt (1st row), blonde hair (2nd row), brown shoulder bag (3rd row) and badge (4th row).  

\paragraph{Comparison with supervised method.} We further provide two well-known supervised methods for reference, including the Part-based Convolutional Baseline (PCB) \cite{sun2018beyond} and the joint Discriminative and Generative Network (DG-Net) \cite{zheng2019joint}. Unsupervised ICE achieves competitive performance with PCB. If we replace the clustering generated pseudo labels with ground truth, our ICE can be transformed into a supervised method. The supervised ICE is competitive with state-of-the-art supervised ReID methods (\eg, DG-Net), which shows that the supervised contrastive learning has a potential to be considered into future supervised ReID. 

\section{Conclusion}
In this paper, we propose a novel inter-instance contrastive encoding method ICE to address unsupervised ReID. Deviated from previous proxy based contrastive ReID methods, we focus on inter-instance affinities to make a model more robust to data variance. We first mine the hardest positive with mini-batch instance pairwise similarity ranking to form a hard instance contrastive loss, which effectively reduces intra-class variance. Smaller intra-class variance contributes to the compactness of clusters. Then, we use mini-batch instance pairwise similarity scores as soft labels to enhance the consistency before and after data augmentation, which makes a model robust to artificial augmentation variance. By combining the proposed hard instance contrastive loss and soft instance consistency loss, ICE significantly outperforms previous unsupervised ReID methods on Market1501, DukeMTMC-reID and MSMT17 datasets. 

\paragraph{Acknowledgements.}
This work has been supported by the French government, through the 3IA Côte d’Azur Investments in the Future project managed by the National Research Agency (ANR) with the reference number ANR-19-P3IA-0002. The authors are grateful to the OPAL infrastructure from Université Côte d'Azur for providing resources and support. 

{\small
\bibliographystyle{ieee_fullname}
\bibliography{egbib}
}


\begin{appendices}
\section{Algorithm Details}  \label{Appendix: Algorithm Details}

The ICE algorithm details are provided in Algorithm~\ref{algo:1}.

\begin{algorithm}[]
    \SetKwInOut{Input}{Input}
    \SetKwInOut{Output}{Output}
    \Input{Unlabeled dataset $\mathcal{X}$, ImageNet pre-trained online encoder $\theta_o$, ImageNet pre-trained momentum encoder $\theta_m$, maximal epoch $E_{max}$ and maximal iteration $I_{max}$.  }
    \Output{Momentum encoder $\theta_m$ after training.}

    \For{$epoch=1$ to $E_{max}$}
      {
        Encode $\mathcal{X}$ to momentum representations $\mathcal{M}$ with the momentum encoder $\theta_m$\;
        Rerank and Generate clustering pseudo labels $\mathcal{Y}$ on momentum representations $\mathcal{M}$ with DBSCAN\;
        Calculate cluster proxies in Eq.~(\ref{equa:center_a}) and camera proxies in Eq.~(\ref{equa:center_c}) based on $\mathcal{Y}$\;
        \For{$iter=1$ to $I_{max}$}
        {
         Calculate inter-instance similarities in a mini-batch\; 
         Train $\theta_o$ with the total loss in Eq.~(\ref{equ:total}) which combines proxy contrastive loss in Eq.~(\ref{equ:proxy}), hard instance contrastive loss in Eq.~(\ref{equa:viewloss}) and soft instance consistency loss in Eq.~(\ref{equa:consistency})\;
         Update $\theta_m$ by Eq.~(\textcolor{red}{\ref{equ:ema}});
        }
      }

    \caption{Inter-instance Contrastive Encoding (ICE) for fully unsupervised ReID.}
    \label{algo:1}
\end{algorithm}

\section{Backbone Network} \label{Appendix: Backbone Network}
Instance-batch normalization (IBN) \cite{pan2018two} has shown better performance than regular batch normalization in unsupervised domain adaptation \cite{pan2018two,ge2020self} and domain generalization \cite{Jia2019FrustratinglyEP}. We compare the performance of ICE with ResNet50 and IBN-ResNet50 backbones in Tab.~\ref{table:backbone}. The performance of our proposed ICE can be further improved with an IBN-ResNet50 backbone network.

\begin{table*}
\scalebox{0.9}{
\begin{tabular}{c|cccc|cccc|cccc}
\hline
\multirow{2}{*}{Backbone}  & \multicolumn{4}{c}{Market1501} & \multicolumn{4}{|c}{DukeMTMC-reID}& \multicolumn{4}{|c}{MSMT17} \\ \cline{2-13}
\multicolumn{1}{c|}{} & \multicolumn{1}{c}{mAP} & \multicolumn{1}{c}{R1} & \multicolumn{1}{c}{R5} & \multicolumn{1}{c|}{R10} & \multicolumn{1}{c}{mAP} & \multicolumn{1}{c}{R1} &\multicolumn{1}{c}{R5} & \multicolumn{1}{c|}{R10} & \multicolumn{1}{c}{mAP} & \multicolumn{1}{c}{R1} &\multicolumn{1}{c}{R5} &\multicolumn{1}{c}{R10} \\ \hline
ResNet50&82.3&93.8&\textbf{97.6}&98.4&69.9&83.3&91.5&94.1&38.9&70.2&80.5&84.4\\
IBN-ResNet50&\textbf{82.5}&\textbf{94.2}&\textbf{97.6}&\textbf{98.5}&\textbf{70.7}&\textbf{83.6}&\textbf{91.9}&\textbf{93.9}&\textbf{40.6}&\textbf{70.7}&\textbf{81.0}&\textbf{84.6}\\
\hline
\end{tabular}}
\centering
\caption{Comparison of ResNet50 and IBN-ResNet50 backbones on Market1501, DukeMTMC-reID and MSMT17 datasets.}
\label{table:backbone}
\end{table*}

\begin{table*}
\scalebox{0.9}{
\begin{tabular}{c|cccc|cccc|cccc}
\hline
\multirow{2}{*}{Threshold}  & \multicolumn{4}{c}{Market1501} & \multicolumn{4}{|c}{DukeMTMC-reID}& \multicolumn{4}{|c}{MSMT17} \\ \cline{2-13}
\multicolumn{1}{c|}{} & \multicolumn{1}{c}{mAP} & \multicolumn{1}{c}{R1} & \multicolumn{1}{c}{R5} & \multicolumn{1}{c|}{R10} & \multicolumn{1}{c}{mAP} & \multicolumn{1}{c}{R1} &\multicolumn{1}{c}{R5} & \multicolumn{1}{c|}{R10} & \multicolumn{1}{c}{mAP} & \multicolumn{1}{c}{R1} &\multicolumn{1}{c}{R5} &\multicolumn{1}{c}{R10} \\ \hline
$0.45$&82.5&93.4&97.5&98.3&68.0&82.8&\textbf{91.5}&93.4&36.6&69.2&79.3&82.7\\
$0.5$&\textbf{83.0}&\textbf{94.1}&\textbf{97.7}&98.3&69.2&82.9&91.2&93.2&38.4&69.9&80.2&83.8\\
$0.55$&82.3&93.8&97.6&98.4&\textbf{69.9}&83.3&\textbf{91.5}&\textbf{94.1}&38.9&70.2&80.5&84.4\\
$0.6$&81.2&93.0&97.3&\textbf{98.5}&69.4&\textbf{83.5}&91.4&94.0&\textbf{39.4}&\textbf{70.9}&\textbf{81.0}&\textbf{84.5}\\
\hline
\end{tabular}}
\centering
\caption{Comparison of different distance thresholds on Market1501, DukeMTMC-reID and MSMT17 datasets.}
\label{table:threshold}
\end{table*}

\section{Threshold in clustering} \label{Appendix: Threshold in clustering}
In DBSCAN \cite{Ester1996ADA}, the distance threshold is the maximum distance between two samples for one to be considered as in the neighborhood of the other. A smaller distance threshold is likely to make DBSCAN mark more hard positives as different classes. On the contrary, a larger distance threshold makes DBSCAN mark more hard negatives as same class.

In the main paper, the distance threshold for DBSCAN between same cluster neighbors is set to $0.55$, which is a trade-off number for Market1501, DukeMTMC-reID and MSMT17 datasets. To get a better understanding of how ICE is sensitive to the distance threshold, we vary the threshold from $0.45$ to $0.6$. As shown in Tab.~\ref{table:threshold}, a smaller threshold $0.5$ is more appreciate for the relatively smaller dataset Market1501, while a larger threshold $0.6$ is more appreciate for the relatively larger dataset MSMT17. State-of-the-art unsupervised ReID methods SpCL \cite{ge2020self} and CAP \cite{Wang2021camawareproxies} respectively used $0.6$ and $0.5$ as their distance threshold. Our proposed ICE can always outperform SpCL and CAP on the three datasets with a threshold between $0.5$ and $0.6$.

\begin{figure}
\centering
   \includegraphics[width=0.9\linewidth]{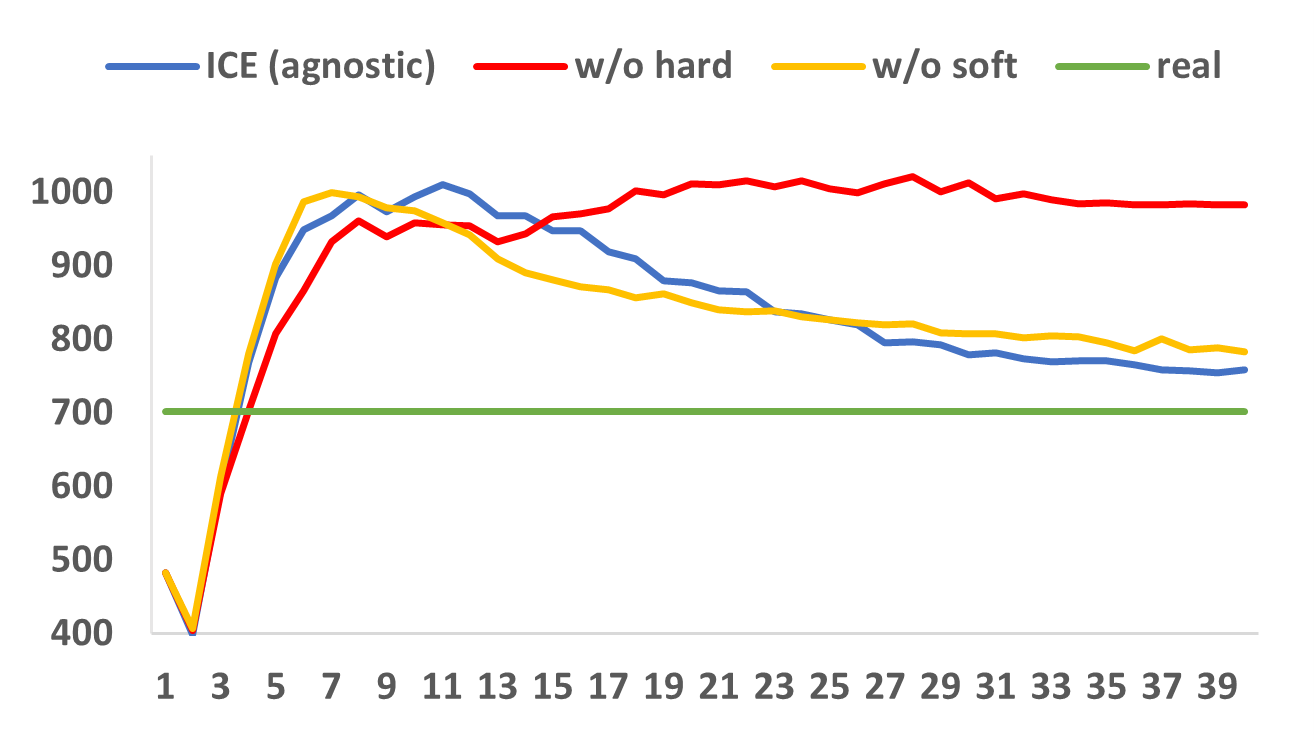}
   \caption{Dynamic cluster numbers of \textbf{ICE(agnostic)} during 40 training epochs on DukeMTMC-reID. A lower number denotes that clusters are more compact (less intra-cluster variance).}
\label{fig:Dynamic cluster numbers agnostic}
\end{figure}

\section{Camera-agnostic scenario} \label{Appendix: Camera-agnostic scenario}
As mentioned in the main paper, we provide the dynamic cluster numbers of camera-agnostic ICE during the training in Fig.~\ref{fig:Dynamic cluster numbers agnostic}. The \textcolor{red}{red curve} is trained without the hard instance contrastive loss $\mathcal{L}_{h\_ins}$ as intra-class variance constraint. In this case, the soft instance consistency loss $\mathcal{L}_{s\_ins}$ maintains high intra-class variance, \eg, $AA\not\approx AP_{1}\not\approx AP_{2}\not\approx AP_{3}\not\approx 1$, which leads to less compact clusters. The \textcolor{orange}{orange curve} is trained without $\mathcal{L}_{s\_ins}$, which has less clusters at the beginning but more clusters at last epochs than the \textcolor{blue}{blue curve}. The \textcolor{blue}{blue curve} is trained with both $\mathcal{L}_{h\_ins}$ and $\mathcal{L}_{s\_ins}$, whose cluster number is most accurate among the three curves at last epochs. Fig.~\ref{fig:Dynamic cluster numbers agnostic} confirms that combining $\mathcal{L}_{h\_ins}$ and $\mathcal{L}_{s\_ins}$ reduces naturally captured and artificially augmented view variance at the same time, which gives optimal ReID performance.

\section{Future work}
Our proposed method is designed for traditional short-term person ReID, in which persons do not change their clothes. 
For long-term person ReID, when persons take off or change their clothes, our method is prone to generate less robust pseudo labels, which relies on visual similarity (mainly based on cloth color). For future work, an interesting direction is to consider how to generate robust pseudo labels to tackle the cloth changing problem for long-term person ReID.

\end{appendices}

\end{document}